\documentclass[10pt,twocolumn,letterpaper]{article}
\usepackage{multirow}
\usepackage{cvpr}
\usepackage{times}
\usepackage{epsfig}
\usepackage{graphicx}
\usepackage{amsmath}
\usepackage{amssymb}
\usepackage{subfig}
\usepackage{xcolor, soul}
\usepackage[pagebackref=true,breaklinks=true,letterpaper=true,colorlinks,bookmarks=false]{hyperref}

 \cvprfinalcopy % *** Uncomment this line for the final submission

\def\cvprPaperID{****} % *** Enter the CVPR Paper ID here

\ifcvprfinal\fi
\begin{document}

%%%%%%%%% TITLE
\title{Beyond Explainability: Leveraging Interpretability for Improved Adversarial Learning}

\author{Devinder Kumar, \text{Ibrahim Ben Daya}$^*$, \text{Kanav Vats}$^*$, \text{Jeffery Feng}, \text{Graham Taylor}$^\dag$, \text{Alexander Wong}\\
University of Waterloo \hspace{2cm}  $^\dag$\text{University of Guelph}\\
Waterloo, Ontario, Canada\hspace{1.5cm} Guelph, Ontario, Canada\\
{\tt\small \{d22kumar,ibendaya,k2vats,j64feng,a28wong\}@uwaterloo.ca} \hspace{0.3cm} \tt\small{gwtaylor@uoguelph.ca}
% For a paper whose authors are all at the same institution,
% omit the following lines up until the closing ``''.
% Additional authors and addresses can be added with ``\and'',
% just like the second author.
% To save space, use either the email address or home page, not both
}

\maketitle
%\thispagestyle{empty}
%%%%%%%%% ABSTRACT
\begin{abstract}
   In this study, we propose the leveraging of interpretability for tasks beyond purely the purpose of explainability. In particular, this study puts forward a novel strategy for leveraging gradient-based interpretability in the realm of adversarial examples, where we use insights gained to aid adversarial learning. More specifically, we introduce the concept of spatially constrained one-pixel adversarial perturbations, where we guide the learning of such adversarial perturbations towards more susceptible areas identified via gradient-based interpretability. Experimental results using different benchmark datasets show that such a spatially constrained one-pixel adversarial perturbation strategy can noticeably improve the speed of convergence as well as produce successful attacks that were also visually difficult to perceive, thus illustrating an effective use of interpretability methods for tasks outside of the purpose of purely explainability. 
\end{abstract}

\let\thefootnote\relax\footnote{*Equal Contribution}
%%%%%%%%% BODY TEXT
\section{Introduction}

In recent times, gradient-based interpretability has grown into a significant area of research in the field of explainable artificial intelligence (XAI). Gradient-based interpretability is generally leveraged to understand data sensitivity~\cite{ gradcam,smoothgrad,zhou2015cnnlocalization}, where the most sensitive areas responsible for a particular prediction made are identified.  In particular, much of previous literature in XAI centers around leveraging gradient-based interpretability for highlighting the regions of interest for any given input to explain the decision making process of deep neural networks.

More recently, there has been some attention in exploring the use of gradient-based interpretability for purposes other than purely explainability.  For example, Zhou et. al.~\cite{zhou2015cnnlocalization} proposed the use of gradient-based interpretability to improve the localization performance of deep neural networks. A number of research studies~\cite{wacv,lee2019ficklenet} have leveraged sensitivity maps produced via gradient-based interpretability as initialization for the task of segmentation.  However, leveraging gradient-based interpretability for tasks beyond explainability is still not well explored outside of these few examples, making further investigations into alternative directions for leveraging insights gained through interpretability ripe for exploration.  

In this study, we investigate and put forward a novel strategy for leveraging gradient-based interpretability in the realm of adversarial examples, where the goal is to produce delicately perturbed inputs designed to mislead machine learning models towards incorrect predictions.  More specifically, we introduce the concept of spatially constrained one-pixel adversarial perturbations, guided by gradient-based interpretability such that insights gained via interpretability is used to aid adversarial learning. One-pixel adversarial perturbations~\cite{onePx} is an extreme case of adversarial examples where only one pixel is modified to fool a model into providing the wrong prediction. This pixel is found through Differential Evolution\cite{DE}, where a population of candidate pixels is randomly modified to create children that compete with its parents for fitness in the next iteration; this fitness criterion being the probabilistic predicted label. The optimal pixels for one-pixel adversarial perturbations usually lie in positions of interest. This observation motivates us to leverage gradient-based interpretability to constrain the differential evolution initialization; we posit that, by ensuring that the initial population of pixels lie in positions of interest as given by generated sensitivity maps, the optimization algorithm for generating one-pixel adversarial perturbations can converge faster with fewer iterations.  Furthermore, by guiding it towards areas of interest, the produced attacks may also be more visually difficult to perceive.
    
    The paper is organized as follows. Section 2 presents the proposed strategy for leveraging gradient-based interpretability in the realm of adversarial examples. Section 3 presents and discusses the experimental results for studying the efficacy of the proposed strategy on several benchmark datasets (CIFAR-10 and NIPS-W). Finally, conclusions are drawn in Section 4.

\section{Methodology}

In this section, we will provide a detailed description of the proposed strategy for leveraging gradient-based interpretability in the realm of adversarial examples, which is designed to improve the speed of convergence for producing adversarial perturbations by using insights gained to aid adversarial learning.

\subsection{Spatially constrained one-pixel adversarial perturbations}

In one-pixel adversarial perturbations, the optimal pixel is found by using Differential Evolution (DE). For every individual adversarial perturbation, a set of $N$ vectors in $\mathbb{R}^5$ - where each vector $p$ represents a candidate pixel's $xy$-coordinates and RGB values - is randomly generated, giving the initial parent population. This initialization serves as input to the differential evolution optimization. For each iteration during optimization, $N$ children are generated from the parent population, and the fittest pixels i.e., the ones providing the lowest probabilistic label for the correct class remain to become $N$ parents in the next iteration.

We hypothesize that the probability of susceptibility is highest in areas that are identified as highly sensitive via gradient-based interpretability. Therefore, we posit that using sensitivity maps generated via gradient-based interpretability as a spatial constraint when generating the initial parent population should speed up DE convergence.

\begin{figure}
\begin{center}
  \subfloat[Susceptibility set generation.]{
		    \includegraphics[width=1\linewidth]{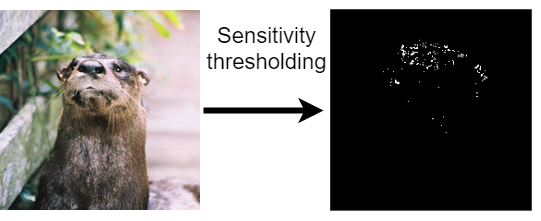}
		   
		}\\
		\subfloat[One-pixel adversarial perturbation generation.]{
		    \includegraphics[width=1\linewidth]{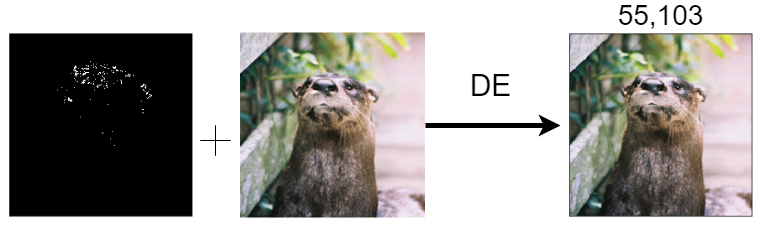}
		} 
\end{center}
  \caption{Pipeline for the generation of spatially constrained one-pixel adversarial perturbations. (a) The set of susceptible pixels is identified via sensitivity thresholding on a sensitivity map obtained using SmoothGrad. (b) The pixel perturbation, identified based on spatial constraint via the set of susceptible pixels, is performed to the image to obtain the adversarial example. The coordinate of the adversarially perturbed pixel is showed on top of the output image. }
\label{fig:random_spatial}
\end{figure}

Fig.~\ref{fig:random_spatial} shows an overview of the pipeline for the generation of spatially constrained one-pixel adversarial perturbations on an image.  

\subsection{Susceptibility set generation}

The first step involves the identification of a set of susceptible pixels $S$ based on insights gained via gradient-based interpretability.  More specifically, we leverage SmoothGrad~\cite{smilkov2017smoothgrad} to obtain a sensitivity map $s$ for each image $x$, with each pixel in $s$ providing a quantitative indicator of importance of the underlying content to the decision making process.  To identify the set of susceptible pixels $S$, binary thresholding is performed on the sensitivity map $s$ as follows: 

\begin{equation}
\begin{aligned}
    s(x) &= \frac{1}{n} \sum_{1}^{n}  \hat s (x+ \mathcal{N}(0,\sigma^2)); \hspace{0.2cm}
    S = \left\{s (x) > \tau\right\}
    \end{aligned}
\end{equation}

This is based on the assumption that the pixels with higher sensitivity in the generated sensitivity maps are more susceptible to attack. 

\subsection{Adversarial perturbation generation}
The second step involves leveraging the set of susceptible pixels for guiding the adversarial learning process. More specifically, when initializing the population for DE, the $X$ and $Y$ values in each vector $p$ for given image $x$ are constrained to be selected from within the set of $S$ i.e., $X,Y \in S$  to ensure that the initial population of pixels all lie in the susceptible regions of the image.

\begin{equation}
\begin{aligned}
    p (x) =\{X,Y,r,g,b\}; \hspace{0.2cm} (X,Y) \in S \\
    \max_{p(x)} f_{adv} (x+p(x)) \hspace{0.1cm}  \vee || p(x)||_{0} \leq 1
    \end{aligned}
\end{equation}
 
\begin{figure}
    \centering
    \includegraphics[width=\linewidth]{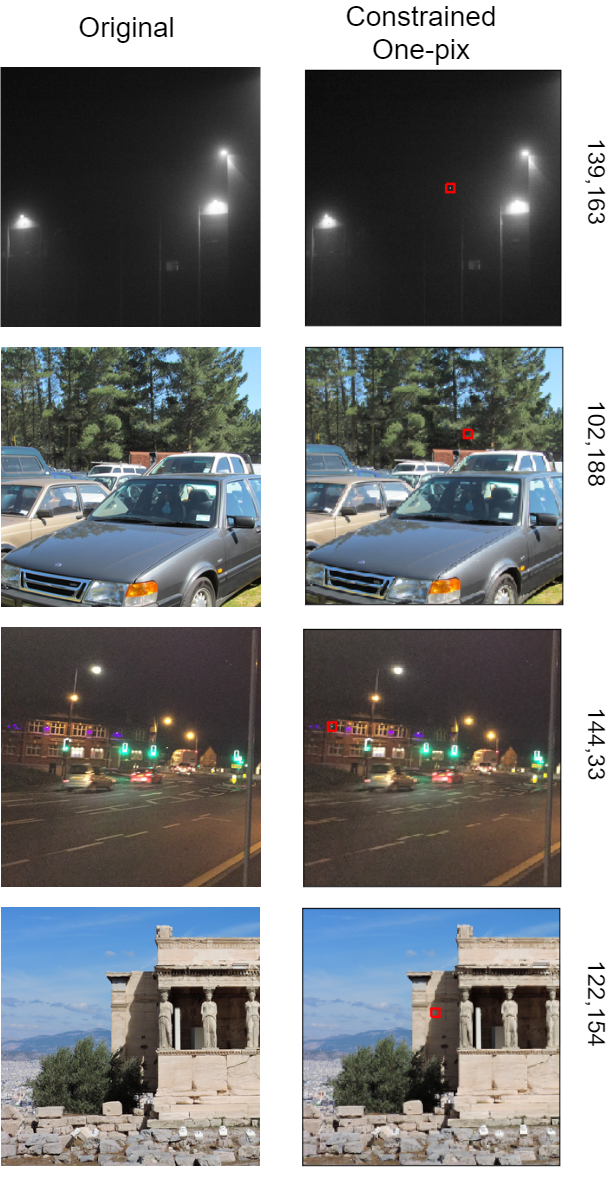}
    \caption{Example images from ImageNet (left) and the corresponding spatially constrained one-pixel adversarial perturbed image (right) that led to successful attacks. The coordinates of the perturbed pixels are shown on the right-hand side. }
    \label{fig:examples_fig}
\end{figure}

\section{Experiments and Results}

To evaluate the efficacy of spatially constrained one-pixel adversarial perturbations in reducing the convergence time of DE, we conducted different experiments on the DE algorithm (both in duration as well as average number of iterations) at finding a susceptible pixel with and without  constrained initialization. Experiments were done for two different datasets: i) CIFAR-10~\cite{krizhevsky}, and ii) NIPS 2017 adversarial attacks and defences challenge dataset~\cite{nips} (small subset derived from ImageNet), referred to as NIPS-W in the rest of the paper.  The network tested in this study was SqueezeNet.

We select the default set of $100$ images taken from the CIFAR-10 dataset in~\cite{onePx}. Additionally, for conducting a larger set of experiments, we randomly select $500$ images from NIPS-W. For all experiments, we set the threshold $\tau$ at 0.5. For each dataset, five different runs of the adversarial attack were performed with different random seeds.

The results for the CIFAR-10 and NIPS-W datasets are summarized in Table~\ref{tab:onepxattack_cifar} and Table~\ref{tab:onepxattack_imagenet} respectively. It can be clearly observed that across different runs and on average, the constrained one-pixel adversarial perturbations converged faster with comparatively less number of iterations when compared to conventional one-pixel adversarial perturbations. This effect is more pronounced for CIFAR-10, with the spatially constrained one-pixel adversarial perturbations converging $6\times$ faster than conventional attacks.  Therefore, it can be clearly observed that by guiding the adversarial learning process for one-pixel adversarial perturbations towards areas of strong susceptibility based on insights gained by gradient-based interpretability, one can accelerate the optimization process to converge in a much more rapid manner.

Example images from ImageNet that were successfully attacked and their corresponding spatially constrained one-pixel adversarial perturbed images are shown in Fig.~\ref{fig:examples_fig}.  It can be observed that the perturbed images, which resulted in successful attacks given that the tested network (SqueezeNet in this study) provided incorrect predictions compared to the original images, look very perceptually similar to the original images with the perturbed pixel visually hidden in most cases. The visual imperceptibility gained when leveraging the proposed spatially constrained one-pixel adversarial perturbation strategy stems from the fact that the adversarial learning process was guided towards sensitivity areas in the images that have complex details that conceal the perturbation well.

\begin{table}[!t]
    \centering
    \caption{Results of constrained one-pixel adversarial perturbations on 100 images from CIFAR-10 dataset. For both constrained and unconstrained perturbations, the total time in seconds, average time in seconds, and the average number of DE iterations are presented. Five runs for constrained and unconstrained one-pixel adversarial perturbations show that constrained perturbations converge faster.}
    \footnotesize
    \setlength{\tabcolsep}{0.2cm}
    \begin{tabular}{c|c|c|c|c|c}\hline
       \multicolumn{3}{c|}{Unconstrained} & \multicolumn{3}{c}{Constrained (ours)}\\ \hline
       T (s) & AVG T (s) & AVG \# itr  & T (s) & AVG T (s) & AVG \# itr \\\hline\hline
       207.98 & 2.08 & 35.61 & 37.15 & 0.56 & 2.52\\ \hline
       216.00 & 2.16 & 37.40 & 35.36 & 0.54 & 2.33\\ \hline
       212.10 & 2.12 & 36.72 & 34.16 & 0.52 & 2.26\\ \hline
       210.71 & 2.11 & 36.39 & 37.03 & 0.58 & 2.52\\ \hline
       210.72 & 2.11 & 36.24 & 36.05 & 0.55 & 2.40\\ \hline
    \end{tabular}
    \label{tab:onepxattack_cifar}
\end{table}

\begin{table}[!t]
    \centering
    \caption{Results of constrained one-pixel adversarial perturbations on the NIPS-W dataset. For both constrained and unconstrained successful attacks, the total time in seconds, average time in seconds, and average number of DE iterations are presented. Five runs for constrained and unconstrained one-pixel adversarial perturbations show that constrained perturbations converge faster.}
    \footnotesize
    \setlength{\tabcolsep}{0.2cm}
    \begin{tabular}{c|c|c|c|c|c}\hline
       \multicolumn{3}{c|}{Constrained (ours)} & \multicolumn{3}{c}{Unconstrained}\\ \hline
       T (s) & AVG T (s) & AVG \# itr  & T (s) & AVG T (s) & AVG \# itr \\\hline\hline
       355.69 & 6.03 & 6.44 & 445.55 & 8.25 &  8.80\\ \hline
       392.21 & 6.88 & 7.63 & 523.31 & 8.58 & 9.70 \\ \hline 
       446.09 & 7.31 & 8.13 & 516.71 & 8.62 & 9.52\\ \hline
       344.90 & 5.95 & 6.45 & 459.81 & 7.79 & 8.75\\ \hline
       416.18 & 7.19 & 7.86 & 559.48 & 9.32 & 10.60\\ \hline
    \end{tabular}
    \label{tab:onepxattack_imagenet}
\end{table}

\section{Conclusion}
In this work, we presented a novel strategy in the realm of adversarial examples that leverage gradient-based interpretability, thus illustrate the use of such methods beyond purely the purpose of explainability. In particular, we introduce a spatially constrained one-pixel adversarial perturbation strategy that leverages identified sensitivity within an image based on gradient-based interpretability, thus leveraging insights gained to aid adversarial learning.  Detailed experiments and evaluations were done to show that leveraging gradient-based interpretability can be used for faster convergence of one-pixel adversarial perturbations. We hope this work will allow for further development of gradient-based interpretability methods to be used for many more tasks and use cases beyond purely explainability. 
%------------------------------------------------------------------------
\section*{Acknowledgment}
This work was partially supported by the Natural Sciences
and Engineering Research Council of Canada and the Canada Research Chairs Program. The authors would also like to thank Nvidia for donating GPUs used in the research.

{\small
\bibliographystyle{ieee}
\bibliography{egbib}
}

\end{document}